\documentclass{article}

\usepackage[preprint]{neurips_2026}

\usepackage[utf8]{inputenc}
\usepackage[T1]{fontenc}
\usepackage{hyperref}
\usepackage{url}
\usepackage{booktabs}
\usepackage{amsmath}
\usepackage{amssymb}
\usepackage{graphicx}
\usepackage{xcolor}
\usepackage{longtable}
\usepackage{array}
\usepackage{calc}
\usepackage{caption}

\DeclareUnicodeCharacter{00B1}{\ensuremath{\pm}}
\DeclareUnicodeCharacter{03C3}{\ensuremath{\sigma}}
\DeclareUnicodeCharacter{0394}{\ensuremath{\Delta}}
\DeclareUnicodeCharacter{2212}{\ensuremath{-}}
\DeclareUnicodeCharacter{2013}{--}
\DeclareUnicodeCharacter{2014}{---}
\DeclareUnicodeCharacter{2192}{\ensuremath{\rightarrow}}

\title{DINO-MVR: Multi-View Readout of Frozen DINOv3 for Annotation-Efficient Medical Segmentation}

\author{
Wei Jiang\\
The University of Queensland\\
\texttt{w.jiang@uq.edu.au}
\And
Feng Liu\\
The University of Queensland\\
\texttt{feng@eecs.uq.edu.au}
\And
Nan Ye\\
The University of Queensland\\
\texttt{nan.ye@uq.edu.au}
\And
Hongfu Sun\\
The University of Newcastle\\
\texttt{hongfu.sun@newcastle.edu.au}
}

\begin{document}

\maketitle

\begin{abstract}
Adapting foundation models to medical segmentation typically requires either backbone fine-tuning or high-capacity task-specific decoders, both of which are difficult to fit reliably when annotations are scarce. We show that frozen DINOv3 features already contain useful structural and boundary cues for medical segmentation, and that the main bottleneck lies in how these features are read out. We propose \textbf{DINO-MVR}, a \emph{Multi-View Readout} framework for annotation-efficient medical segmentation. DINO-MVR trains only lightweight MLP probes on features from the final three transformer blocks of a frozen DINOv3 backbone, without updating the backbone itself. At inference, each input is interpreted through complementary resolutions and test-time augmentations, whose probability maps are combined by entropy-weighted fusion and refined with simple spatial regularization. For volumetric inputs, Gaussian z-axis smoothing further improves inter-slice consistency. Under fixed evaluation protocols on endoscopy, dermoscopy, and MRI benchmarks, DINO-MVR achieves strong readout-only performance, including 0.895 Dice on Kvasir-SEG, 0.897 Dice on ISIC 2018, and 0.908 Dice on BraTS FLAIR whole-tumor segmentation. With only five annotated BraTS patients, it recovers 98.4\% of the performance obtained by the 40-patient BraTS reference run. These results suggest that frozen self-supervised vision backbones can support accurate medical segmentation when paired with an effective multi-view readout.
\end{abstract}

\section{Introduction}\label{introduction}

Medical image segmentation is central to clinical and biomedical image analysis,
including lesion quantification, organ delineation, treatment planning, and
longitudinal follow-up. Supervised systems such as U-Net and nnU-Net remain
strong baselines when sufficient in-domain labels are available
\citep{ronneberger2015unet,isensee2021nnu}. Yet pixel- and voxel-level
annotations are costly, slow, and often require specialist expertise. Moreover,
models trained on one dataset can degrade under changes in scanner, acquisition
protocol, contrast, institution, anatomy, or target structure, making repeated
annotation and adaptation difficult to avoid.

Vision foundation models offer a possible way to reduce this dependence on dense
medical labels. Promptable models such as SAM and MedSAM show that large-scale
pretraining can produce reusable segmentation capabilities
\citep{kirillov2023segment,ma2024medsam}. Self-supervised transformers from the
DINO family provide a complementary route: they learn patch-level visual
representations without segmentation labels, and DINOv3 reports strong dense
feature quality across visual tasks without task-specific fine-tuning
\citep{caron2021emerging,oquab2023dinov2,simeoni2025dinov3}. This raises a
focused question for medical segmentation: if the backbone is kept completely
frozen, how much segmentation performance can be recovered by improving only the
readout?

We study this question in a deliberately constrained setting. Given an
off-the-shelf DINOv3 ViT-B/16 backbone, we freeze all encoder parameters and
train only lightweight task-specific probes. This separates two issues that are
often optimized together: whether the pretrained representation contains useful
medical structure, and how effectively a downstream module converts that
structure into a dense mask. Recent frozen-DINO segmentation work such as
SegDINO suggests that lightweight readouts can be competitive
\citep{yang2025segdino}; we take the next step by treating readout itself as the
main design space. A frozen representation can be queried along several axes,
including feature depth, input resolution, and test-time transformations.
Relying on a single readout view may expose only part of the available feature
evidence. We therefore ask whether multiple complementary views of the same
representation can be combined to obtain stronger annotation-efficient
segmentation.

We propose \textbf{DINO-MVR}, a \emph{Multi-View Readout} framework for frozen
DINOv3 medical segmentation. During training, DINO-MVR fits small MLP probes on
patch features from the last three transformer blocks, allowing the readout to
access complementary late-layer features without updating the backbone. During
inference, it queries the frozen representation through multiple input
resolutions and test-time augmentations. It then aggregates the resulting
probability maps with entropy-guided selective fusion and DenseCRF refinement
\citep{krahenbuhl2011efficient}.
For volumetric inputs, DINO-MVR applies the same slice-wise readout and adds a
simple Gaussian smoothing prior along the z-axis to improve inter-slice
consistency without learning a 3D decoder.

We report results under fixed evaluation protocols on Kvasir-SEG, ISIC 2018,
and BraTS FLAIR \citep{jha2020kvasir,isic2018,baid2021rsna}.
DINO-MVR achieves strong readout-only performance, including 0.895 Dice on
Kvasir-SEG, 0.897 Dice on ISIC 2018, and 0.908 Dice on BraTS FLAIR whole-tumor
segmentation, while updating only lightweight probes. In a
patient-limited BraTS setting, five annotated patients recover 98.4\% of the
40-patient reference performance. These results suggest that frozen
self-supervised dense features can support accurate medical segmentation when
paired with an effective multi-view readout.

Our contributions are threefold:
\begin{enumerate}
\item We formulate frozen-DINOv3 medical segmentation as a readout problem,
isolating how much dense prediction performance can be extracted without
backbone adaptation.
\item We introduce DINO-MVR, which combines feature-depth, resolution, and
test-time transformation views using selective fusion and lightweight spatial
refinement.
\item We demonstrate strong 2D and volumetric performance under fixed protocols,
including annotation-efficient adaptation from a small number of labeled
patients and a non-parametric volumetric consistency prior.
\end{enumerate}

\section{Related Work}\label{related-work}

\subsection{Supervised and annotation-efficient medical segmentation}
\label{supervised-and-annotation-efficient-medical-segmentation}

Medical segmentation is commonly built on supervised encoder--decoder networks.
U-Net established the skip-connected architecture that remains a standard design
for biomedical dense prediction, while nnU-Net showed that preprocessing,
augmentation, training configuration, ensembling, and post-processing are also
crucial to strong performance \citep{ronneberger2015unet,isensee2021nnu}. These
systems are powerful, but adapting them to new modalities or target structures
often requires new dense annotations and retraining.

Few-shot medical image segmentation reduces this burden by using annotated
support examples to segment query images or volumes. Representative methods
include prototype-based SSL-ALPNet \citep{ouyang2020self}, attention- or
transformer-enhanced matching methods such as CRAPNet and RPT
\citep{ding2023few,zhu2023few}, and recent approaches that improve prototype
quality or extend the support--query formulation to 3D data
\citep{fan2025spenet,zheng2024few}. DINO-MVR targets the same annotation
bottleneck but uses a different protocol: lightweight probes are trained once on
a small labeled set, after which inference is fully automatic and does not rely
on support annotations or prompts for each test case.

\subsection{Foundation models, adaptation, and frozen-feature readout}
\label{foundation-models-adaptation-and-frozen-feature-readout}

Segmentation foundation models shift the field from task-specific training
toward reusable pretrained models. SAM introduced promptable segmentation at
large natural-image scale, and MedSAM adapts this paradigm to medical images
with large-scale medical image-mask supervision
\citep{kirillov2023segment,ma2024medsam}. Other medical foundation models build
stronger domain-specific encoders or universal segmentation systems, such as
UNETR for transformer-based 3D segmentation and SuPreM for supervised 3D medical
pretraining \citep{hatamizadeh2022unetr,li2024supervised}. Parameter-efficient
adaptation further reduces the cost of modifying large models through prompt
parameters or adapters \citep{jia2022visual,fischer2024prompt,wu2025medicalsamadapter}.
These lines are complementary to our work, but they typically require prompts,
medical-domain pretraining, or some form of model adaptation. DINO-MVR instead
keeps the natural-image DINOv3 backbone fixed and confines task learning to
lightweight readout probes.

Frozen-feature probing provides a way to measure what information is accessible
in a pretrained representation without end-to-end adaptation
\citep{alain2016understanding}. DINO-style self-supervised ViTs are especially
relevant for dense probing: DINO revealed emergent object-level structure,
DINOv2 showed broad transfer from large self-supervised features, and DINOv3
emphasizes high-quality dense representations
\citep{caron2021emerging,oquab2023dinov2,simeoni2025dinov3}. The closest
frozen-DINO segmentation baseline is SegDINO, which pairs a frozen DINOv3
backbone with a lightweight segmentation decoder/readout for medical and natural
image segmentation \citep{yang2025segdino}. DINO-MVR builds on this direction
but focuses less on a single decoder head and more on how to query and fuse
multiple views of a fixed feature space.

\subsection{Multi-view inference and uncertainty-aware fusion}
\label{multi-view-inference-and-uncertainty-aware-fusion}

Multi-scale inference, test-time augmentation, ensembling, and post-processing
are widely used to improve segmentation robustness. nnU-Net incorporates many of
these choices in a self-configuring pipeline \citep{isensee2021nnu}, while
medical segmentation work has used test-time augmentation to characterize
transformation-induced uncertainty \citep{wang2019aleatoric}. DenseCRF-style
post-processing can further sharpen predictions by encouraging spatially and
appearance-consistent labels \citep{krahenbuhl2011efficient}.

DINO-MVR uses these tools in a specific frozen-representation setting. Its views
are not merely a generic ensemble over image transforms: late-layer feature views
expose complementary states of the final DINOv3 representation, resolution views
trade global context against local detail, and test-time transformation views
test spatial stability. The resulting predictions are combined by
entropy-guided selective fusion, with simple spatial and z-axis regularization
added only at inference time. This positions DINO-MVR as a structured readout
strategy for frozen self-supervised features, complementary to promptable
foundation models and parameter-efficient adaptation.

\section{Method}
\label{method}

\begin{figure}[t]
\centering
\captionsetup{font=footnotesize}
\includegraphics[width=\linewidth]{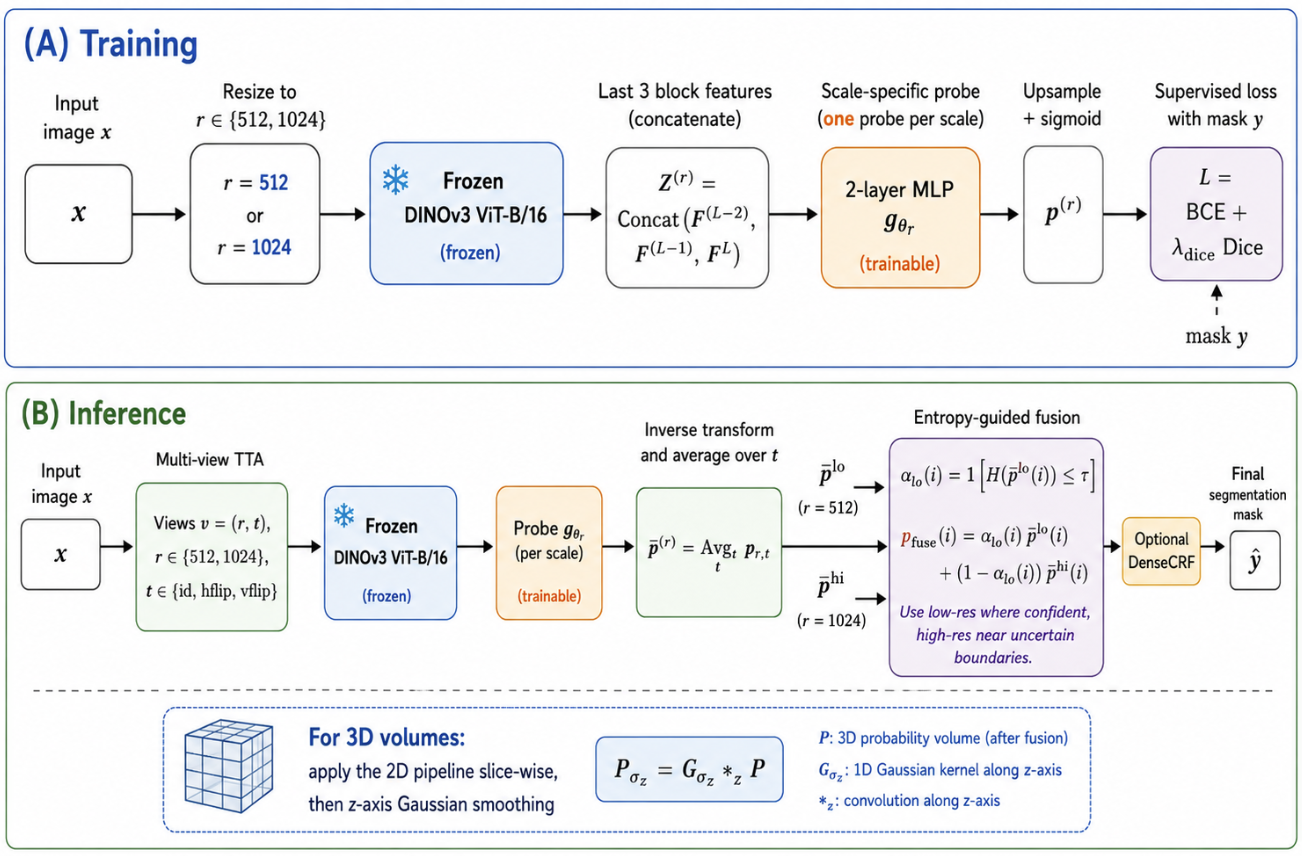}
\caption{DINO-MVR method overview. Frozen DINOv3 features are read out by
scale-specific MLP probes and fused across test-time and resolution views.}
\label{fig:method_overview}
\end{figure}

DINO-MVR is a deliberately simple readout protocol for medical segmentation with
an off-the-shelf self-supervised vision backbone. Rather than introducing a new
segmentation backbone or a high-capacity decoder, we keep the DINOv3 encoder
fixed and restrict task learning to lightweight MLP probes. The method has two
stages. During training, scale-specific probes learn to map frozen patch features
to foreground probabilities. During inference, each image is evaluated through
multiple complementary views---input resolution and test-time transformation---
and the resulting probability maps are fused using a selective routing rule,
followed by optional non-parametric spatial refinement. Figure~\ref{fig:method_overview}
summarizes this pipeline.

\subsection{Problem setup}
\label{sec:problem_setup}

Let $x \in \mathbb{R}^{H \times W \times C}$ denote an input image and
$y \in \{0,1\}^{H \times W}$ its binary segmentation mask. We focus on the
common foreground--background setting used in our benchmarks; the same readout
can be extended to multi-class segmentation by replacing the final sigmoid with a
softmax over classes. Let $F_{\psi}$ be a pretrained DINOv3 ViT-B/16 encoder
\citep{simeoni2025dinov3}. Throughout training and inference, the backbone
parameters $\psi$ are frozen. Only the probe parameters are optimized. This
constraint isolates the central question of this work: how much medical
segmentation performance can be recovered from information already present in the
frozen representation?

\subsection{Frozen feature stack}
\label{sec:frozen_feature_stack}

For an input resized to resolution $r$, denoted $x^{(r)}$, the frozen ViT
produces a grid of patch-token features. For transformer layer $\ell$, with $L$
denoting the final transformer block, we write
\begin{equation}
  F_{\psi}^{\ell}(x^{(r)}) \in \mathbb{R}^{h_r \times w_r \times d},
  \qquad h_r = w_r = r/16,
  \label{eq:layer_features}
\end{equation}
where $d=768$ for ViT-B/16. Following the intuition that nearby final layers
encode complementary levels of local structure and semantic context, we use the
last $N=3$ transformer blocks and concatenate their patch features:
\begin{equation}
  Z^{(r)}(x) = \operatorname{Concat}\!\big(
    F_{\psi}^{L-2}(x^{(r)}),
    F_{\psi}^{L-1}(x^{(r)}),
    F_{\psi}^{L}(x^{(r)})
  \big)
  \in \mathbb{R}^{h_r \times w_r \times 3d}.
  \label{eq:feature_stack}
\end{equation}
For single-channel medical images such as MRI, CT, and ultrasound, the input is
replicated across three channels before applying the standard DINOv3
preprocessing. Because $F_{\psi}$ is frozen, gradients are never propagated
through the backbone; training memory and task-specific capacity are therefore
concentrated in the readout probes.

\subsection{Lightweight scale-specific readout}
\label{sec:lightweight_readout}

DINO-MVR trains one two-layer MLP probe per resolution. For a resolution
$r \in \mathcal{R}$, the probe
$g_{\theta_r}:\mathbb{R}^{3d}\rightarrow\mathbb{R}$ maps each concatenated patch
feature to a foreground logit:
\begin{equation}
  a^{(r)} = g_{\theta_r}\!\big(Z^{(r)}(x)\big),
  \label{eq:patch_logits}
\end{equation}
where $a^{(r)} \in \mathbb{R}^{h_r \times w_r}$ is a patch-level logit map. In
our implementation, $g_{\theta_r}$ has architecture $2304 \rightarrow 256
\rightarrow 1$ with a ReLU hidden layer, corresponding to approximately $0.59$M
trainable parameters per scale. The patch logits are bilinearly upsampled to the
mask resolution and converted to foreground probabilities:
\begin{equation}
  p^{(r)} = \operatorname{sigmoid}\!\left(\operatorname{Upsample}(a^{(r)})\right),
  \label{eq:scale_probability}
\end{equation}
where $\operatorname{sigmoid}$ is applied element-wise.

The probes are trained with a standard combination of binary cross-entropy and
soft Dice loss \citep{milletari2016vnet}:
\begin{equation}
  \mathcal{L}
  = \mathcal{L}_{\mathrm{BCE}}(p,y)
  + \lambda_{\mathrm{dice}}\mathcal{L}_{\mathrm{Dice}}(p,y), \qquad
  \mathcal{L}_{\mathrm{Dice}}
  = 1 - \frac{2\sum_i p_i y_i + \epsilon}{\sum_i p_i + \sum_i y_i + \epsilon}.
  \label{eq:training_loss}
\end{equation}
Here $\lambda_{\mathrm{dice}}$ controls the contribution of the Dice term. Only
$\{\theta_r\}_{r\in\mathcal{R}}$ are updated. When augmentation is enabled, we
use lightweight geometric transformations such as flips, rotations, and scale
changes. These transformations are applied to the input image and mask before
feature extraction, so they do not modify the frozen backbone.

\subsection{Multi-view inference}
\label{sec:multi_view_inference}

At inference time, DINO-MVR evaluates each image under a small set of views. A
base readout view is defined by an input resolution $r$ and a test-time
transformation $t$:
\begin{equation}
  v=(r,t), \qquad r\in\mathcal{R}, \quad t\in\mathcal{T}.
  \label{eq:view_definition}
\end{equation}
In our experiments, $\mathcal{R}=\{512,1024\}$ and
$\mathcal{T}=\{\mathrm{id},\mathrm{hflip},\mathrm{vflip}\}$. For each view, the
transformed image is encoded by the frozen backbone, decoded by the corresponding
scale-specific probe, inverse-transformed, and resized back to the original
image coordinates:
\begin{equation}
  p_{r,t}
  = t^{-1}\!\left[
    \operatorname{sigmoid}\!\left(
      \operatorname{Upsample}
      \left(g_{\theta_r}\!\left(Z^{(r)}(t(x))\right)\right)
    \right)
  \right].
  \label{eq:view_prediction}
\end{equation}
The predictions over test-time transformations are averaged within each
resolution:
\begin{equation}
  \bar{p}^{(r)} = \frac{1}{|\mathcal{T}|}
  \sum_{t\in\mathcal{T}} p_{r,t}.
  \label{eq:tta_average}
\end{equation}
This training-free step reduces orientation-specific variation and is consistent
with the use of test-time augmentation for uncertainty estimation in medical
segmentation \citep{wang2019aleatoric}.

\subsection{Resolution fusion and spatial refinement}
\label{sec:resolution_fusion}

The two resolutions provide complementary evidence. The lower-resolution branch
is typically more stable in large homogeneous regions, while the higher-resolution
branch preserves finer boundary detail. We combine them with an entropy-guided
hard weighting rule. For a probability $p$ at pixel $i$, binary entropy is
\begin{equation}
  \mathcal{H}(p_i) = -p_i\log(p_i+\epsilon) - (1-p_i)\log(1-p_i+\epsilon).
  \label{eq:binary_entropy}
\end{equation}
Let $\bar{p}^{\mathrm{lo}}$ and $\bar{p}^{\mathrm{hi}}$ denote the TTA-averaged
predictions from the $512$ and $1024$ branches. The fused prediction is
\begin{equation}
  p_{\mathrm{fuse}}(i)
  = \alpha_{\mathrm{lo}}(i)\,\bar{p}^{\mathrm{lo}}(i)
  + \big(1-\alpha_{\mathrm{lo}}(i)\big)\,\bar{p}^{\mathrm{hi}}(i),
  \qquad
  \alpha_{\mathrm{lo}}(i)=
  \mathbf{1}\!\big[\mathcal{H}(\bar{p}^{\mathrm{lo}}(i)) \leq \tau\big].
  \label{eq:entropy_guided_fusion}
\end{equation}
We use $\tau=0.3$ unless otherwise specified. This rule uses the coarse branch
where it is confident and falls back to the high-resolution branch in uncertain
regions, especially around object boundaries. It introduces no additional
trainable parameters.

After fusion, we optionally apply DenseCRF refinement to encourage spatially and
appearance-consistent masks \citep{krahenbuhl2011efficient}. The CRF uses
standard Gaussian and bilateral pairwise terms and is applied only at inference
time; no CRF parameters are learned. The final binary mask for 2D inputs is
obtained by thresholding the refined probability map at $0.5$.

\subsection{Volumetric extension}
\label{sec:volumetric_extension}

For volumetric inputs, we apply the same 2D readout pipeline slice by slice and
then impose a simple non-parametric consistency prior along the depth axis. Let
$P(q,z)$ denote the fused probability at in-plane pixel $q$ for slice $z$. We
smooth the volume with a one-dimensional Gaussian kernel along $z$ only:
\begin{equation}
  P_{\sigma_z}(q,z)
  = \left(G_{\sigma_z} *_z P\right)(q,z),
  \label{eq:z_smoothing}
\end{equation}
where $*_z$ denotes convolution along the slice dimension. We use $\sigma_z=4$ in
volumetric experiments. This step suppresses isolated slice-wise errors and
improves inter-slice consistency without training a 3D decoder or modifying the
frozen DINOv3 backbone. The final volumetric mask is obtained by thresholding
$P_{\sigma_z}$ at $0.5$.

\section{Experiments}
\label{sec:experiments}

We organize the evaluation around the two main claims of DINO-MVR. First,
\emph{annotation-efficient adaptation}: because the backbone already provides
reusable dense features, a small number of annotated medical cases can be
sufficient to train the lightweight probes. Second, \emph{multi-view readout}:
when the DINOv3 backbone is frozen, reading the same representation through
complementary views can recover stronger segmentation masks than a single fixed
readout. All experiments keep the DINOv3 encoder frozen; only the MLP readouts
are optimized.

Unless otherwise stated, Dice similarity coefficient (DSC) is the primary
metric. We additionally report intersection-over-union (IoU) and the
95th-percentile Hausdorff distance (HD95) when available. For volumetric
datasets, metrics are computed after reconstructing the 3D prediction from
slice-wise probability maps.

\subsection{Experimental setup}
\label{sec:experimental_setup}

\paragraph{Datasets.}
The fixed-protocol evidence comes from three benchmarks with
distinct roles. Kvasir-SEG is used as the 2D endoscopy setting for multi-view
readout and component ablation. ISIC 2018 provides a second 2D setting in
dermoscopy, using 800 training images and a held-out 600-image test export with
no overlap with the saved probe-training split. BraTS 2021 FLAIR is used as the
volumetric MRI setting for patient-limited annotation-efficiency experiments.

\paragraph{Implementation.}
All experiments use the same frozen DINOv3 ViT-B/16 feature extractor. For each
image or slice, we concatenate patch tokens from the final three transformer
blocks and train a two-layer MLP probe, $2304\!\rightarrow\!256\!\rightarrow\!1$,
to predict foreground logits. The backbone is never updated. A single
scale-specific probe has approximately $0.59$M trainable parameters. The default
2D pipeline uses probes at $512$ and $1024$ input resolutions, averages identity,
horizontal-flip, and vertical-flip test-time views, fuses the resolution-specific
probability maps with entropy-guided routing, optionally applies DenseCRF
refinement, and thresholds the final map at $0.5$. For volumetric inputs, we
apply the same pipeline slice by slice and smooth the probability volume along
the z-axis with a Gaussian kernel; unless otherwise noted, volumetric
experiments use $\sigma_z=4$.

\paragraph{Baselines and comparison scope.}
For 2D automatic segmentation, we include the published Kvasir-SEG and ISIC
2018 numbers reported by SegDINO \citep{yang2025segdino} as literature context.
In that benchmark, all compared 2D methods are evaluated on images resized to
$256\times256$, and SegDINO uses a DINOv3-S backbone. Our 2D runs use a
DINOv3-B backbone with readout probes at $512$ and $1024$ input resolutions; the
reported 2D metrics are computed after resizing predictions and masks to the
same $256\times256$ metric grid. These numbers are therefore still not strict
head-to-head comparisons, because exact splits, metric implementations,
backbone sizes, and model-selection rules can differ across papers. Promptable
methods that require oracle points or boxes are excluded from the automatic
segmentation setting because prompts change the task definition. For volumetric
experiments, we focus on the internal BraTS K-patient protocol rather than
mixing in external few-shot protocols with incompatible shot definitions.

\subsection{Multi-view readout of frozen DINOv3 features}
\label{sec:multiview_readout}

A frozen DINOv3 backbone exposes several complementary sources of evidence:
feature depth, input resolution, and test-time transformations. We use
Kvasir-SEG to isolate these components under a
matched protocol, while also reporting cross-dataset baselines for scale.

\paragraph{Contextual comparison.}
Table~\ref{tab:benchmark_context}
places DINO-MVR alongside published automatic segmentation results across
Kvasir-SEG and ISIC 2018, reporting DSC, IoU, and HD95. Because HD95 is
pixel-scale dependent, the 2D HD95 entries should be read as resolution-specific
context rather than a matched-resolution ranking. For Kvasir-SEG and ISIC,
Table~\ref{tab:benchmark_context} reports the DenseCRF-refined TTA output; HD95
is averaged over finite boundary-distance cases, with empty-mask failure rates
reported in the table note. BraTS is reported separately in the
annotation-efficiency subsection because it uses patient-limited volumetric
adaptation rather than 2D image-level training.

\begin{table}[t]
\centering
\caption{Contextual comparison on 2D automatic medical image segmentation.
Baseline values are paper-reported by SegDINO \citep{yang2025segdino} at
$256\times256$ resolution.
DINO-MVR uses $512{+}1024$ readout resolutions, but reports 2D metrics on the
same $256\times256$ metric grid. Matched-split reruns would be required for
strict ranking.}
\label{tab:benchmark_context}
\resizebox{\linewidth}{!}{%
\begin{tabular}{lcccccc}
\toprule
Method & \multicolumn{3}{c}{Kvasir-SEG} & \multicolumn{3}{c}{ISIC 2018} \\
\cmidrule(lr){2-4}\cmidrule(lr){5-7}
 & DSC$\uparrow$ & IoU$\uparrow$ & HD95$\downarrow$ & DSC$\uparrow$ & IoU$\uparrow$ & HD95$\downarrow$ \\
\midrule
U-Net~\citep{ronneberger2015unet} & 0.7916 & 0.7029 & 41.58 & 0.8187 & 0.7295 & 25.12 \\
SegNet~\citep{badrinarayanan2017segnet} & 0.8415 & 0.7565 & 25.89 & 0.8327 & 0.7446 & 21.41 \\
R2U-Net~\citep{alom2018recurrent} & 0.7367 & 0.6328 & 45.64 & 0.8102 & 0.7134 & 25.36 \\
Att-UNet~\citep{oktay2018attention} & 0.8016 & 0.7202 & 31.92 & 0.8275 & 0.7372 & 26.12 \\
TransUNet~\citep{chen2021transunet} & 0.8054 & 0.7093 & 37.67 & 0.8186 & 0.7230 & 24.76 \\
U-NeXt~\citep{valanarasu2022unext} & 0.6271 & 0.5164 & 58.32 & 0.8230 & 0.7327 & 23.63 \\
U-KAN~\citep{li2025ukan} & 0.7217 & 0.6381 & 36.92 & 0.8341 & 0.7462 & 23.57 \\
SegDINO~\citep{yang2025segdino} & 0.8765 & 0.8064 & 20.80 & 0.8576 & 0.7760 & 17.80 \\
\midrule
DINO-MVR$^\dagger$ & \textbf{0.8946} & \textbf{0.8375} & \textbf{15.00} & \textbf{0.8976} & \textbf{0.8270} & \textbf{12.40} \\
\bottomrule
\end{tabular}}

\vspace{0.35em}
\footnotesize{$^\dagger$Dice and IoU are computed over the full test export.
HD95 is averaged over cases with finite boundary distance, with undefined
infinite HD95 values counted separately. For Kvasir-SEG, the DenseCRF-refined
TTA output has 3/200 empty-mask HD95 failures (1.50\%). For ISIC 2018, it has
1/600 such failure (0.17\%).}
\end{table}

\paragraph{Ablations.}
A full-pipeline comparison cannot by itself isolate which readout choices matter.
We therefore run matched Kvasir-SEG ablations under the same frozen DINOv3
ViT-B/16 backbone, 800/200 split, $0.5$ threshold, $256\times256$ metric grid,
and finite-case HD95 convention. Table~\ref{tab:mvr_ablation} separates two
questions. Subtable (a) tests the learned readout design by retraining probes
when feature depth or probe type changes. Subtable (b) keeps the saved
last-three MLP probes fixed and removes inference-time views.

\begin{table}[t]
\centering
\caption{Kvasir-SEG ablations under matched protocols.}
\label{tab:mvr_ablation}

\textbf{(a) Readout design ablation.}
\vspace{0.25em}

\resizebox{\linewidth}{!}{%
\begin{tabular}{llccccc}
\toprule
Configuration & Changed factor & DSC$\uparrow$ & $\Delta$DSC & IoU$\uparrow$ & HD95$_{\mathrm{finite}}\downarrow$ & Inf HD95 \\
\midrule
Full DINO-MVR & Last 3 blocks, MLP, $512{+}1024$ & \textbf{0.8946} & 0.0000 & \textbf{0.8375} & 15.0 & 3/200 \\
Last-1 MLP MVR & feature depth & 0.8915 & -0.0031 & 0.8344 & 14.7 & 4/200 \\
Last-2 MLP MVR & feature depth & 0.8909 & -0.0037 & 0.8342 & \textbf{14.0} & 4/200 \\
\midrule
Linear last-3 MVR & probe capacity & 0.8635 & -0.0311 & 0.7941 & 22.8 & 2/200 \\
$512$ only + TTA + DenseCRF & resolution view & 0.8885 & -0.0061 & 0.8261 & 17.7 & 3/200 \\
\bottomrule
\end{tabular}}

\vspace{0.75em}

\textbf{(b) Inference-time view ablation with fixed saved probes.}
\vspace{0.25em}

\resizebox{\linewidth}{!}{%
\begin{tabular}{llccccc}
\toprule
Configuration & Changed factor & DSC$\uparrow$ & $\Delta$DSC & IoU$\uparrow$ & HD95$_{\mathrm{finite}}\downarrow$ & Inf HD95 \\
\midrule
Full DINO-MVR & none & 0.8946 & 0.0000 & 0.8375 & \textbf{15.0} & 3/200 \\
w/o DenseCRF & boundary refinement & \textbf{0.8975} & +0.0029 & \textbf{0.8429} & 17.1 & 3/200 \\
w/o flip-based TTA & test-time views & 0.8875 & -0.0071 & 0.8260 & 20.7 & 2/200 \\
$1024$ only + TTA + DenseCRF & resolution view & 0.8865 & -0.0081 & 0.8290 & 18.2 & 2/200 \\
w/o TTA and DenseCRF & two inference views & 0.8832 & -0.0114 & 0.8193 & 23.4 & 2/200 \\
\midrule
Raw $1024$ readout only & single raw branch & 0.8663 & -0.0283 & 0.7983 & 31.1 & 0/200 \\
Raw $512$ readout only & single raw branch & 0.8644 & -0.0302 & 0.7893 & 26.8 & 1/200 \\
\bottomrule
\end{tabular}}

\vspace{0.35em}
\footnotesize{HD95$_{\mathrm{finite}}$ excludes cases with undefined infinite
boundary distance; the number of such cases is reported explicitly in the final
column. $\Delta$DSC is measured relative to the full DINO-MVR row within each
subtable. Raw single-branch rows in (b) are supporting controls rather than
strict leave-one-out removals from the full pipeline.}
\end{table}

\begin{figure}[t]
\centering
\includegraphics[width=\linewidth]{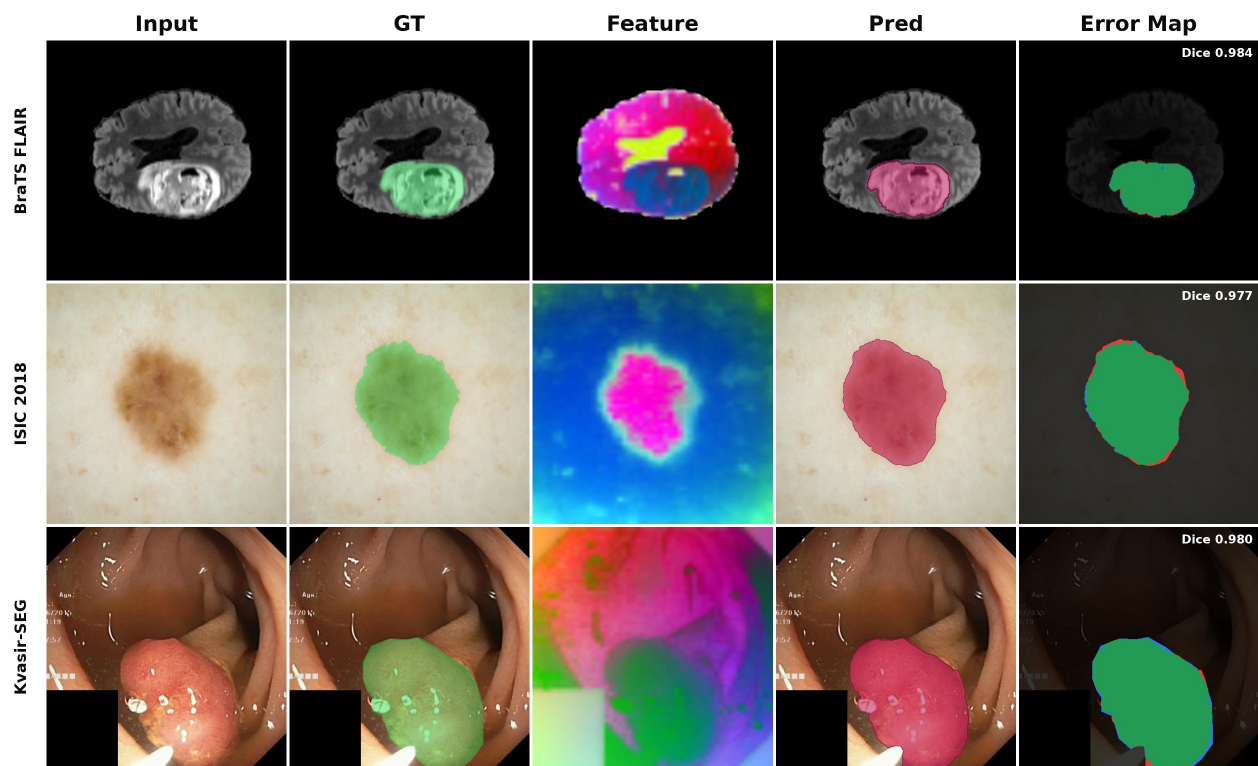}
\caption{Qualitative examples from held-out BraTS FLAIR, ISIC 2018, and
Kvasir-SEG samples. Columns show the input image or slice, ground truth,
PCA visualization of frozen DINO features, DINO-MVR prediction, and error map.
In the error map, green denotes overlap between prediction and ground truth,
red denotes false positives, and blue denotes false negatives.}
\label{fig:qualitative_examples}
\end{figure}

The ablation supports three design choices. First, concatenating the last three
transformer blocks gives the best overlap score, although the one- and two-block
readouts remain close, suggesting that the final DINOv3 blocks already contain
strong dense cues. Second, replacing the MLP with a linear probe causes the
largest drop, indicating that a small nonlinear readout is important even when
the backbone is frozen. Third, using only the $512$ branch weakens both Dice and
HD95 relative to the default $512{+}1024$ readout, supporting the
multi-resolution part of DINO-MVR. The fixed-probe ablation further shows that
flip-based TTA is important for overlap, while DenseCRF behaves mainly as a
boundary regularizer: removing it slightly increases Dice (0.8975) but worsens
finite HD95 (17.1 versus 15.0). We therefore keep DenseCRF in the reported
default pipeline. Figure~\ref{fig:qualitative_examples} shows representative
held-out examples across the three evaluated imaging domains.

\subsection{Annotation-efficient adaptation from few patients}
\label{sec:annotation_efficiency}

We evaluate annotation efficiency on BraTS FLAIR whole-tumor segmentation. We
sample $K$ labeled patients from a fixed training split, use all annotated active
slices from those patients to fit the slice-wise MLP readouts, and evaluate on
held-out patients. The reference run uses 40 labeled patients from the
same fixed split under the same frozen-backbone pipeline.

\begin{figure}[t]
\centering
\includegraphics[width=\linewidth]{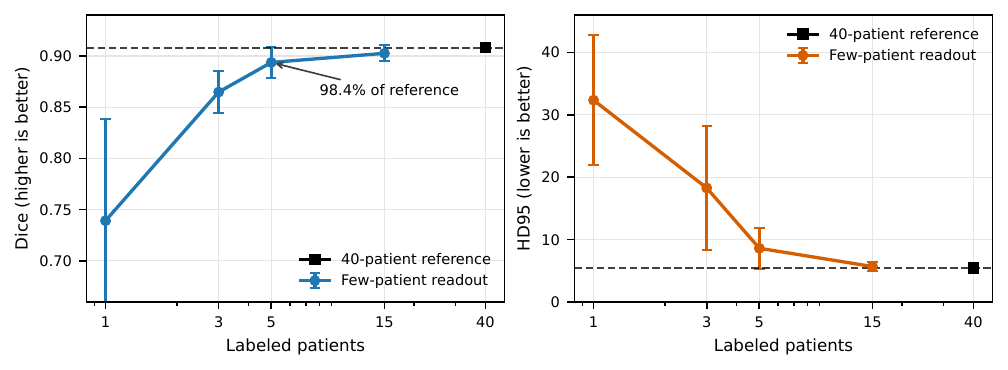}
\caption{BraTS FLAIR K-patient learning curve. Points show mean performance over
patient-sampling seeds, with error bars denoting one standard deviation when
multiple seeds are available. The square marker and dashed line denote the
40-patient reference readout.}
\label{fig:brats_k_patient}
\end{figure}

Figure~\ref{fig:brats_k_patient} shows that DINO-MVR adapts quickly as
annotations are added. With five annotated patients,
it reaches $0.8937$ DSC, recovering $98.4\%$ of the 40-patient reference
performance ($0.9082$ DSC) while using only $12.5\%$ of the labeled patients.
The largest improvement occurs between $K=1$ and $K=5$, after which the curve
begins to saturate. Boundary quality follows the same trend: HD95 decreases from
$32.34$ at $K=1$ to $8.59$ at $K=5$ and $5.42$ in the 40-patient reference run.
These results support annotation-efficient adaptation: once the backbone is
frozen, the amount of task-specific learning needed to obtain a usable
volumetric segmenter can be small. Because adaptation only trains the MLP
readouts, the same protocol takes 92.3 seconds with five annotated patients,
compared with 675.6 seconds for the 40-patient reference readout.

\paragraph{Volumetric consistency without a learned 3D decoder.}
Table~\ref{tab:z_smoothing} evaluates the z-axis smoothing prior on a BraTS FLAIR
pilot run with $K=10$ patients and seed 42. Smoothing improves DSC over the
slice-wise baseline and reduces HD95 substantially. The best DSC occurs at
$\sigma_z=4$, while $\sigma_z=3$--$5$ gives similar Dice and lower HD95 than no
smoothing. This supports the design choice of adding a simple non-parametric
volumetric prior rather than training a high-capacity 3D decoder.

\begin{table}[t]
\centering
\caption{BraTS z-axis smoothing pilot ablation.}
\label{tab:z_smoothing}
\begin{tabular}{ccccc}
\toprule
z-smooth $\sigma_z$ & DSC & IoU & HD95 & $\Delta$ DSC \\
\midrule
0 (none) & 0.8958 & 0.8135 & 11.9 & baseline \\
2 & 0.9041 & 0.8267 & 8.9 & +0.0083 \\
3 & 0.9056 & 0.8292 & 6.9 & +0.0098 \\
4 & \textbf{0.9057} & 0.8291 & 6.2 & +0.0099 \\
5 & 0.9043 & 0.8268 & \textbf{5.9} & +0.0085 \\
\bottomrule
\end{tabular}
\end{table}

\section{Discussion and Limitations}
\label{discussion-and-limitations}

DINO-MVR studies a deliberately constrained question: if a DINOv3 encoder is kept
fixed, how much medical segmentation performance can be recovered by changing
only the readout? The experiments support two observations. First, combining
feature-depth, resolution, and test-time transformation views can extract strong
automatic 2D masks from frozen features. Second, in BraTS FLAIR, few-patient
probes recover most of the 40-patient reference performance, suggesting that a
substantial part of task adaptation can be shifted from backbone training to
lightweight readout fitting.

\paragraph{Scope of the evidence.}
The fixed-protocol evidence covers Kvasir-SEG, ISIC 2018, and BraTS
FLAIR. The 2D comparisons to SegDINO-reported baselines are best interpreted as
positioning rather than strict ranking, because published baselines may differ in
split construction, input resolution, metric implementation, and model-selection
rules. The BraTS K-patient study uses an internal patient-level protocol and is
intended to measure annotation-efficient adaptation rather than test-time
conditioning on annotated support examples. A strict comparison would require
rerunning baselines under the same automatic, prompt-free protocol.

\paragraph{Efficiency and inference cost.}
DINO-MVR is parameter-efficient in the adaptation sense: only small MLP probes
are trained. This should not be conflated with end-to-end computational
efficiency. Full inference includes frozen DINOv3 feature extraction at multiple
resolutions, flip-based test-time aggregation, entropy-guided fusion, DenseCRF
refinement, and optional z-axis smoothing for volumes. These choices are useful
for testing the multi-view readout hypothesis, but their runtime and memory costs
should be reported separately from trainable-parameter counts.

\paragraph{Choice of views and volumetric modeling.}
The multi-view framing does not imply that every additional view is useful.
Preliminary wavelet and three-resolution variants did not provide sufficiently
consistent gains to include in the default protocol. For 3D inputs, DINO-MVR
uses slice-wise 2D readouts plus a non-parametric z-axis smoothing prior; this is
simple and annotation-efficient, but it cannot replace learned 3D models when
long-range volumetric context or anatomy-specific shape constraints are
essential.

\section{Conclusion}
\label{conclusion}

We introduced DINO-MVR, a frozen-backbone framework that treats medical
segmentation as multi-view readout over DINOv3 features. The method trains only
lightweight scale-specific MLP probes, combines predictions from multiple
resolutions and test-time views, and adds simple spatial regularization. The
fixed-protocol results suggest that frozen self-supervised vision features can
serve as a practical starting point for medical segmentation when paired with an
effective readout. Broader modality generalization and fully matched comparisons
to task-specific segmentation models remain important next steps.

\clearpage
\bibliographystyle{plainnat}
\bibliography{references}

@inproceedings{ronneberger2015unet,
  title     = {U-Net: Convolutional Networks for Biomedical Image Segmentation},
  author    = {Ronneberger, Olaf and Fischer, Philipp and Brox, Thomas},
  booktitle = {Medical Image Computing and Computer-Assisted Intervention -- MICCAI 2015},
  pages     = {234--241},
  year      = {2015},
  publisher = {Springer}
}

@article{badrinarayanan2017segnet,
  title   = {{SegNet}: A Deep Convolutional Encoder-Decoder Architecture for Image Segmentation},
  author  = {Badrinarayanan, Vijay and Kendall, Alex and Cipolla, Roberto},
  journal = {IEEE Transactions on Pattern Analysis and Machine Intelligence},
  volume  = {39},
  number  = {12},
  pages   = {2481--2495},
  year    = {2017}
}

@misc{alom2018recurrent,
  title         = {Recurrent Residual Convolutional Neural Network Based on {U-Net} ({R2U-Net}) for Medical Image Segmentation},
  author        = {Alom, Md Zahangir and Hasan, Mahmudul and Yakopcic, Chris and Taha, Tarek M. and Asari, Vijayan K.},
  year          = {2018},
  eprint        = {1802.06955},
  archivePrefix = {arXiv},
  primaryClass  = {cs.CV}
}

@misc{oktay2018attention,
  title         = {Attention {U-Net}: Learning Where to Look for the Pancreas},
  author        = {Oktay, Ozan and Schlemper, Jo and Le Folgoc, Loic and Lee, Matthew and Heinrich, Mattias and Misawa, Kazunari and Mori, Kensaku and McDonagh, Steven and Hammerla, Nils Y. and Kainz, Bernhard and Glocker, Ben and Rueckert, Daniel},
  year          = {2018},
  eprint        = {1804.03999},
  archivePrefix = {arXiv},
  primaryClass  = {cs.CV}
}

@misc{chen2021transunet,
  title         = {{TransUNet}: Transformers Make Strong Encoders for Medical Image Segmentation},
  author        = {Chen, Jieneng and Lu, Yongyi and Yu, Qihang and Luo, Xiangde and Adeli, Ehsan and Wang, Yan and Lu, Le and Yuille, Alan L. and Zhou, Yuyin},
  year          = {2021},
  eprint        = {2102.04306},
  archivePrefix = {arXiv},
  primaryClass  = {cs.CV}
}

@inproceedings{valanarasu2022unext,
  title     = {{UNeXt}: {MLP}-based Rapid Medical Image Segmentation Network},
  author    = {Valanarasu, Jeya Maria Jose and Patel, Vishal M.},
  booktitle = {Medical Image Computing and Computer Assisted Intervention -- MICCAI 2022},
  pages     = {23--33},
  year      = {2022},
  publisher = {Springer}
}

@inproceedings{li2025ukan,
  title     = {{U-KAN} Makes Strong Backbone for Medical Image Segmentation and Generation},
  author    = {Li, Chenxin and Liu, Xinyu and Li, Wuyang and Wang, Cheng and Liu, Hengyu and Liu, Yifan and Chen, Zhen and Yuan, Yixuan},
  booktitle = {Proceedings of the AAAI Conference on Artificial Intelligence},
  volume    = {39},
  pages     = {4652--4660},
  year      = {2025}
}

@article{isensee2021nnu,
  title   = {nnU-Net: a self-configuring method for deep learning-based biomedical image segmentation},
  author  = {Isensee, Fabian and Jaeger, Paul F. and Kohl, Simon A. A. and Petersen, Jens and Maier-Hein, Klaus H.},
  journal = {Nature Methods},
  volume  = {18},
  number  = {2},
  pages   = {203--211},
  year    = {2021}
}

@inproceedings{kirillov2023segment,
  title     = {Segment Anything},
  author    = {Kirillov, Alexander and Mintun, Eric and Ravi, Nikhila and Mao, Hanzi and Rolland, Chloe and Gustafson, Laura and Xiao, Tete and Whitehead, Spencer and Berg, Alexander C. and Lo, Wan-Yen and Doll{\'a}r, Piotr and Girshick, Ross},
  booktitle = {Proceedings of the IEEE/CVF International Conference on Computer Vision},
  pages     = {4015--4026},
  year      = {2023}
}

@article{ma2024medsam,
  title   = {Segment anything in medical images},
  author  = {Ma, Jun and He, Yuting and Li, Feifei and Han, Lin and You, Chenyu and Wang, Bo},
  journal = {Nature Communications},
  volume  = {15},
  number  = {1},
  pages   = {654},
  year    = {2024}
}

@inproceedings{caron2021emerging,
  title     = {Emerging Properties in Self-Supervised Vision Transformers},
  author    = {Caron, Mathilde and Touvron, Hugo and Misra, Ishan and J{\'e}gou, Herv{\'e} and Mairal, Julien and Bojanowski, Piotr and Joulin, Armand},
  booktitle = {Proceedings of the IEEE/CVF International Conference on Computer Vision},
  pages     = {9650--9660},
  year      = {2021}
}

@misc{oquab2023dinov2,
  title         = {DINOv2: Learning Robust Visual Features without Supervision},
  author        = {Oquab, Maxime and Darcet, Timoth{\'e}e and Moutakanni, Th{\'e}o and Vo, Huy and Szafraniec, Marc and Khalidov, Vasil and Fernandez, Pierre and Haziza, Daniel and Massa, Francisco and El-Nouby, Alaaeldin and Assran, Mahmoud and Ballas, Nicolas and Galuba, Wojciech and Howes, Russell and Huang, Po-Yao and Li, Shang-Wen and Misra, Ishan and Rabbat, Michael and Sharma, Vasu and Synnaeve, Gabriel and Xu, Hu and J{\'e}gou, Herv{\'e} and Mairal, Julien and Labatut, Patrick and Joulin, Armand and Bojanowski, Piotr},
  year          = {2023},
  eprint        = {2304.07193},
  archivePrefix = {arXiv},
  primaryClass  = {cs.CV}
}

@misc{simeoni2025dinov3,
  title         = {DINOv3},
  author        = {Sim{\'e}oni, Oriane and Vo, Huy V. and Seitzer, Maximilian and Baldassarre, Federico and Oquab, Maxime and Jose, Cijo and Khalidov, Vasil and Szafraniec, Marc and Yi, Seungeun and Ramamonjisoa, Micha{\"e}l and Massa, Francisco and Haziza, Daniel and Wehrstedt, Luca and Wang, Jianyuan and Darcet, Timoth{\'e}e and Moutakanni, Th{\'e}o and Sentana, Leonel and Roberts, Claire and Vedaldi, Andrea and Tolan, Jamie and Brandt, John and Couprie, Camille and Mairal, Julien and J{\'e}gou, Herv{\'e} and Labatut, Patrick and Bojanowski, Piotr},
  year          = {2025},
  eprint        = {2508.10104},
  archivePrefix = {arXiv},
  primaryClass  = {cs.CV}
}

@misc{yang2025segdino,
  title         = {SegDINO: An Efficient Design for Medical and Natural Image Segmentation with DINO-V3},
  author        = {Yang, Sicheng and Wang, Hongqiu and Xing, Zhaohu and Chen, Sixiang and Zhu, Lei},
  year          = {2025},
  eprint        = {2509.00833},
  archivePrefix = {arXiv},
  primaryClass  = {cs.CV}
}

@inproceedings{krahenbuhl2011efficient,
  title     = {Efficient Inference in Fully Connected CRFs with Gaussian Edge Potentials},
  author    = {Kr{\"a}henb{\"u}hl, Philipp and Koltun, Vladlen},
  booktitle = {Advances in Neural Information Processing Systems},
  volume    = {24},
  year      = {2011}
}

@inproceedings{ouyang2020self,
  title     = {Self-Supervision with Superpixels: Training Few-shot Medical Image Segmentation without Annotation},
  author    = {Ouyang, Cheng and Biffi, Carlo and Chen, Chen and Kart, Turkay and Qiu, Huaqi and Rueckert, Daniel},
  booktitle = {Computer Vision -- ECCV 2020},
  pages     = {762--780},
  year      = {2020},
  publisher = {Springer}
}

@inproceedings{ding2023few,
  title     = {Few-shot Medical Image Segmentation with Cycle-resemblance Attention},
  author    = {Ding, Hao and Sun, Changchang and Tang, Hao and Cai, Dawen and Yan, Yan},
  booktitle = {Proceedings of the IEEE/CVF Winter Conference on Applications of Computer Vision},
  pages     = {2488--2497},
  year      = {2023}
}

@inproceedings{zhu2023few,
  title     = {Few-Shot Medical Image Segmentation via a Region-Enhanced Prototypical Transformer},
  author    = {Zhu, Yazhou and Wang, Shidong and Xin, Tong and Zhang, Haofeng},
  booktitle = {Medical Image Computing and Computer Assisted Intervention -- MICCAI 2023},
  pages     = {271--280},
  year      = {2023},
  publisher = {Springer}
}

@inproceedings{fan2025spenet,
  title     = {{SPENet}: Self-guided Prototype Enhancement Network for Few-shot Medical Image Segmentation},
  author    = {Fan, Chao and Jia, Xibin and Xiao, Anqi and Yu, Hongyuan and Yang, Zhenghan and Yang, Dawei and Xu, Hui and Huang, Yan and Wang, Liang},
  booktitle = {Medical Image Computing and Computer Assisted Intervention -- MICCAI 2025},
  pages     = {584--593},
  year      = {2025},
  publisher = {Springer Nature Switzerland}
}

@inproceedings{zheng2024few,
  title     = {Few-Shot 3D Volumetric Segmentation with Multi-Surrogate Fusion},
  author    = {Zheng, Meng and Planche, Benjamin and Gao, Zhongpai and Chen, Terrence and Radke, Richard J. and Wu, Ziyan},
  booktitle = {Medical Image Computing and Computer Assisted Intervention -- MICCAI 2024},
  pages     = {286--296},
  year      = {2024},
  publisher = {Springer Nature Switzerland}
}

@inproceedings{hatamizadeh2022unetr,
  title     = {{UNETR}: Transformers for 3D Medical Image Segmentation},
  author    = {Hatamizadeh, Ali and Tang, Yucheng and Nath, Vishwesh and Yang, Dong and Myronenko, Andriy and Landman, Bennett and Roth, Holger R. and Xu, Daguang},
  booktitle = {2022 IEEE/CVF Winter Conference on Applications of Computer Vision (WACV)},
  pages     = {1748--1758},
  year      = {2022},
  doi       = {10.1109/WACV51458.2022.00181}
}

@inproceedings{li2024supervised,
  title     = {How Well Do Supervised 3D Models Transfer to Medical Imaging Tasks?},
  author    = {Li, Wenxuan and Yuille, Alan and Zhou, Zongwei},
  booktitle = {International Conference on Learning Representations},
  year      = {2024}
}

@inproceedings{jia2022visual,
  title     = {Visual Prompt Tuning},
  author    = {Jia, Menglin and Tang, Luming and Chen, Bor-Chun and Cardie, Claire and Belongie, Serge and Hariharan, Bharath and Lim, Ser-Nam},
  booktitle = {Computer Vision -- ECCV 2022},
  pages     = {709--727},
  year      = {2022},
  publisher = {Springer Nature Switzerland}
}

@article{fischer2024prompt,
  title   = {Prompt tuning for parameter-efficient medical image segmentation},
  author  = {Fischer, Marc and Bartler, Alexander and Yang, Bin},
  journal = {Medical Image Analysis},
  volume  = {91},
  pages   = {103024},
  year    = {2024},
  doi     = {10.1016/j.media.2023.103024}
}

@article{wu2025medicalsamadapter,
  title   = {Medical {SAM} adapter: Adapting segment anything model for medical image segmentation},
  author  = {Wu, Junde and Wang, Ziyue and Hong, Mingxuan and Ji, Wei and Fu, Huazhu and Xu, Yanwu and Xu, Min and Jin, Yueming},
  journal = {Medical Image Analysis},
  volume  = {102},
  pages   = {103547},
  year    = {2025},
  doi     = {10.1016/j.media.2025.103547}
}

@misc{alain2016understanding,
  title         = {Understanding intermediate layers using linear classifier probes},
  author        = {Alain, Guillaume and Bengio, Yoshua},
  year          = {2016},
  eprint        = {1610.01644},
  archivePrefix = {arXiv},
  primaryClass  = {stat.ML}
}

@article{wang2019aleatoric,
  title   = {Aleatoric uncertainty estimation with test-time augmentation for medical image segmentation with convolutional neural networks},
  author  = {Wang, Guotai and Li, Wenqi and Aertsen, Michael and Deprest, Jan and Ourselin, S{\'e}bastien and Vercauteren, Tom},
  journal = {Neurocomputing},
  volume  = {338},
  pages   = {34--45},
  year    = {2019},
  doi     = {10.1016/j.neucom.2019.01.103}
}

@inproceedings{milletari2016vnet,
  title        = {{V-Net}: Fully Convolutional Neural Networks for Volumetric Medical Image Segmentation},
  author       = {Milletari, Fausto and Navab, Nassir and Ahmadi, Seyed-Ahmad},
  booktitle    = {2016 Fourth International Conference on 3D Vision (3DV)},
  pages        = {565--571},
  year         = {2016},
  organization = {IEEE},
  doi          = {10.1109/3DV.2016.79}
}

@inproceedings{jha2020kvasir,
  title     = {Kvasir-SEG: A Segmented Polyp Dataset},
  author    = {Jha, Debesh and Smedsrud, Pia H. and Riegler, Michael A. and Halvorsen, P{\aa}l and de Lange, Thomas and Johansen, Dag and Johansen, H{\aa}vard D.},
  booktitle = {MultiMedia Modeling},
  pages     = {451--462},
  year      = {2020},
  publisher = {Springer}
}

@misc{isic2018,
  title = {{ISIC} 2018: Skin Lesion Analysis Towards Melanoma Detection},
  author = {{International Skin Imaging Collaboration}},
  year = {2018},
  note = {Challenge dataset}
}

@misc{baid2021rsna,
  title         = {The RSNA-ASNR-MICCAI BraTS 2021 Benchmark on Brain Tumor Segmentation and Radiogenomic Classification},
  author        = {Baid, Ujjwal and Ghodasara, Satyajeet and Mohan, Suyash and Bilello, Michel and Calabrese, Evan and Colak, Errol and Farahani, Keyvan and Kalpathy-Cramer, Jayashree and Kitamura, Felipe C. and Pati, Spyridon and Prevedello, Luciano M. and Rudie, Jeffrey D. and Sako, Christian and Shinohara, Russell T. and Bergquist, Timothy and Chai, Rong and Eddy, James and Elliott, Jack and Reade, Walter and Schaffter, Thomas and Yu, Tong and Zheng, Jacob and Moawad, Abdulhaleem W. and Coelho, Luiz O. and McDonnell, Olivia and Miller, Elaine and Moron, Florencia E. and Oswood, Marisa C. and Shih, George and Siakallis, Leonidas and Bronstein, Yury and Mason, Jessica and Miller, Angela F. and Choudhury, Adityo and Agarwal, Alok and Besada, Carlos and Derakhshan, Jamshid and Diogo, Mariana C. and Do-Dai, Duong and Farage, Lana and Go, John L. and Jahanipour, Jamshid and Jones, Derrick T. and Kotsenas, Amy and Schneider, Tyler and Linguraru, Marius George and Mongan, John and Shah, Lubdha M. and Ugga, Lorenzo and Bakas, Spyridon},
  year          = {2021},
  eprint        = {2107.02314},
  archivePrefix = {arXiv},
  primaryClass  = {eess.IV}
}

\clearpage
\appendix
\section{Additional qualitative examples}
\label{app:additional_qualitative}

\begin{figure}[p]
\centering
\includegraphics[width=\linewidth]{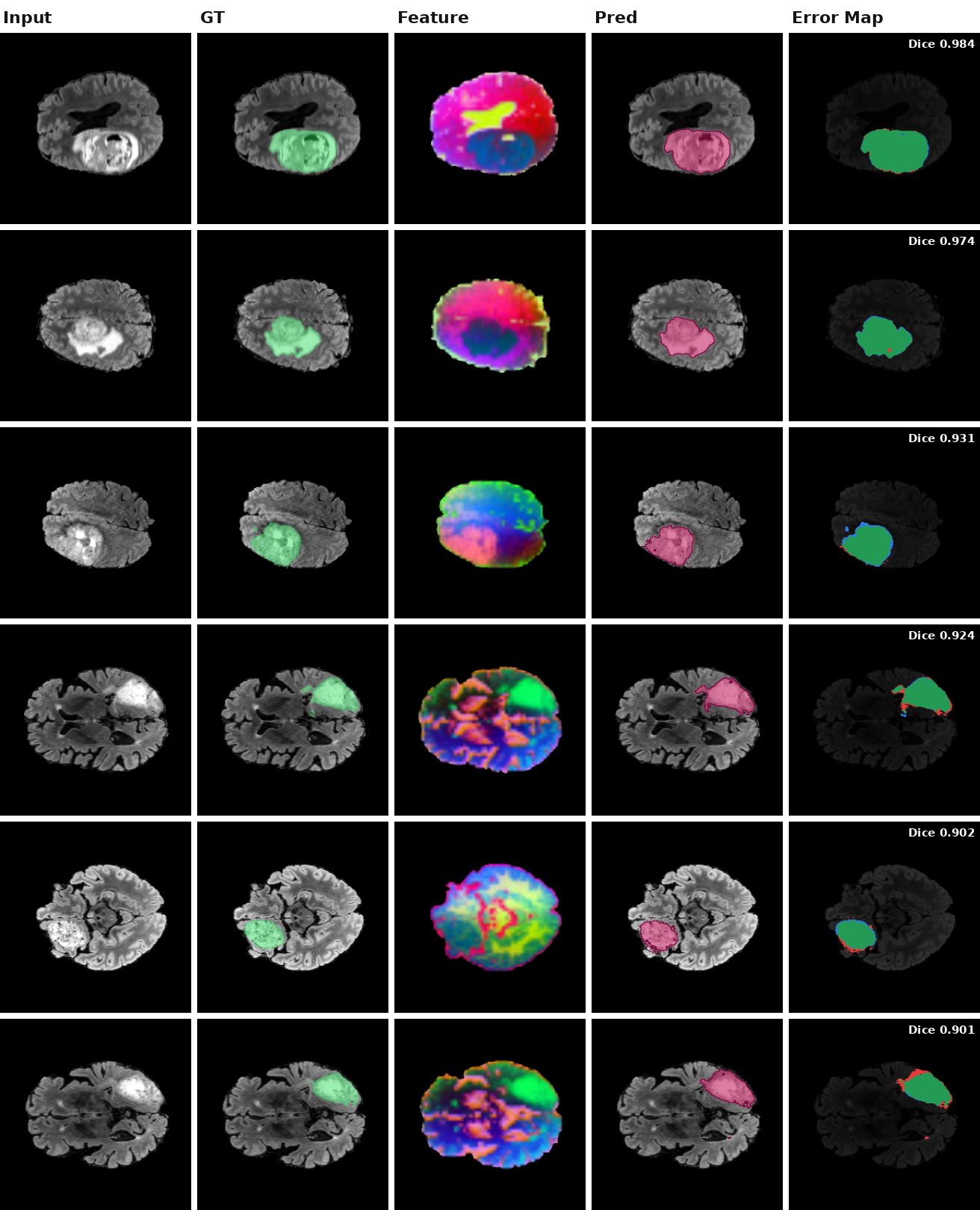}
\caption{Additional BraTS FLAIR qualitative examples. Rows are ordered from
high-scoring to medium-low cases.}
\label{fig:appendix_qual_brats}
\end{figure}

\begin{figure}[p]
\centering
\includegraphics[width=\linewidth]{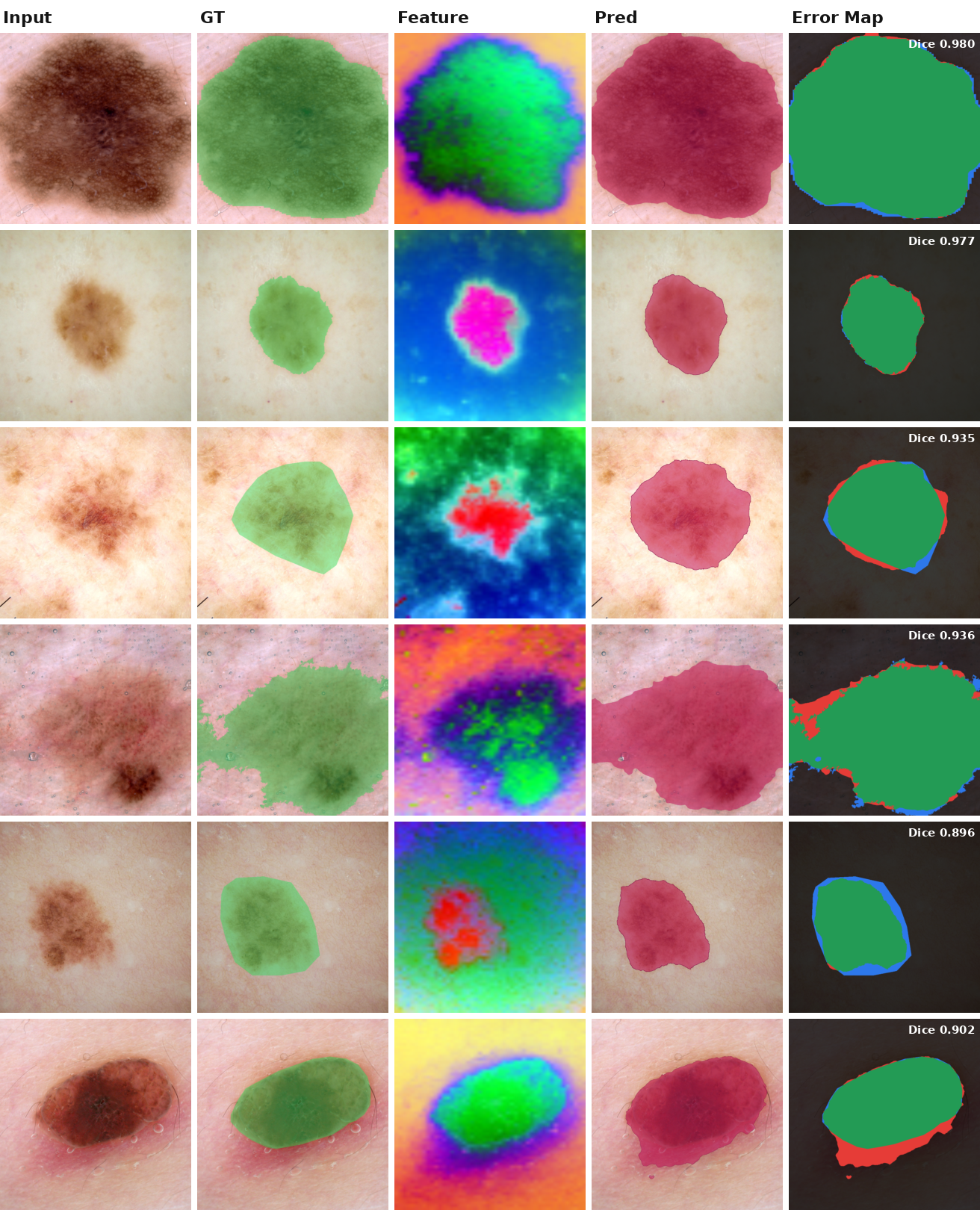}
\caption{Additional ISIC 2018 qualitative examples. Rows are ordered from
high-scoring to medium-low cases.}
\label{fig:appendix_qual_isic}
\end{figure}

\begin{figure}[p]
\centering
\includegraphics[width=\linewidth]{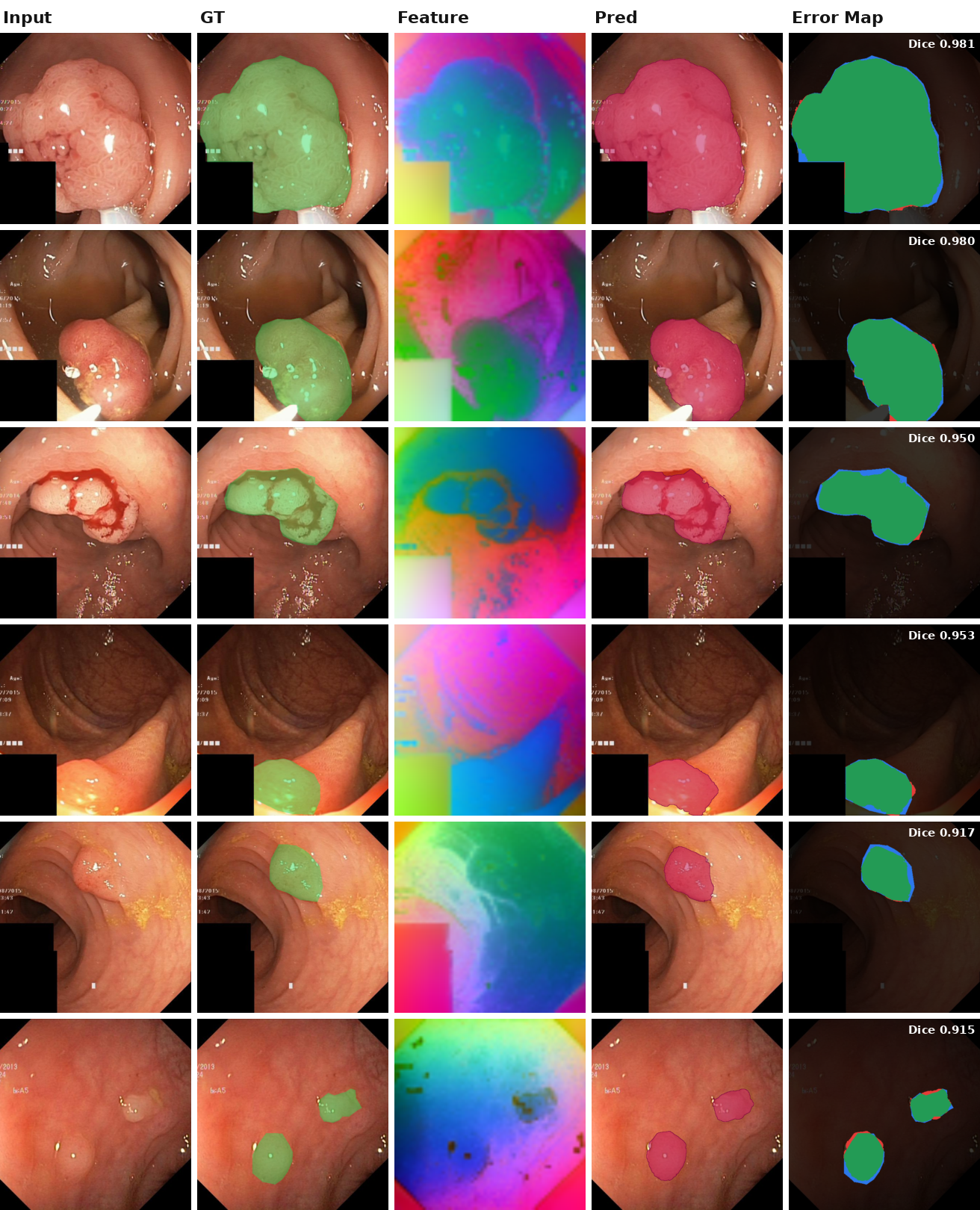}
\caption{Additional Kvasir-SEG qualitative examples. Rows are ordered from
high-scoring to medium-low cases.}
\label{fig:appendix_qual_kvasir}
\end{figure}

Figures~\ref{fig:appendix_qual_brats}, \ref{fig:appendix_qual_isic}, and
\ref{fig:appendix_qual_kvasir}
supplement Figure~\ref{fig:qualitative_examples} with additional held-out
examples from each evaluated domain. For each dataset, we select two high-scoring
cases, two medium cases, and two medium-low cases from the 50-case qualitative
bank used to construct Figure~\ref{fig:qualitative_examples}. The columns match
the main qualitative figure: input, ground truth, frozen-feature visualization,
prediction, and error map. Per-case Dice is shown in the upper-right corner of
the error map.

\end{document}